\documentclass[conference]{IEEEtran}
\IEEEoverridecommandlockouts
\usepackage{cite}
\usepackage{amsmath,amssymb,amsfonts}
\usepackage{graphicx}
\usepackage{textcomp}
\usepackage{xcolor}

\usepackage{algorithm}
\usepackage[noend]{algpseudocode}

\def\BibTeX{{\rm B\kern-.05em{\sc i\kern-.025em b}\kern-.08em
    T\kern-.1667em\lower.7ex\hbox{E}\kern-.125emX}}

\usepackage{tabularx}
\usepackage{longtable}
\usepackage{threeparttable} 
\usepackage{etoolbox} 
\appto\TPTnoteSettings{\footnotesize} 
\usepackage{multirow}
\usepackage{adjustbox}

\usepackage{booktabs}
\usepackage[fleqn]{mathtools} 
\usepackage{bm} 
\usepackage{mathrsfs} 
\usepackage{bbm}
\usepackage{tikz}
\usetikzlibrary{shapes,snakes} 
\usetikzlibrary{arrows.meta}
\newcounter{counter}

\definecolor{GRAY}{gray}{0.97}

\usepackage{hyperref} 
\hypersetup{
    breaklinks = true,
    hyperindex = true,
    colorlinks=false,
    citebordercolor = white,
    filebordercolor = white,
    linkbordercolor = white,
    menubordercolor = white,
    urlbordercolor = white,
    runbordercolor = white,
}

\newcommand\ceil[1]{\lceil #1 \rceil}

\newcommand\bfdiv[2]{\displaystyle\left\lfloor\dfrac{#1}{#2}\right\rfloor}
\newcommand\bcdiv[2]{\displaystyle\left\lceil \dfrac{#1}{#2}\right\rceil}
\usepackage{ragged2e}

\allowdisplaybreaks

\begin{document}

\title{
    State-of-the-art predictive and prescriptive analytics for IEEE CIS 3\textsuperscript{rd} Technical Challenge\\
{\footnotesize \textsuperscript{} }
}

\author{\IEEEauthorblockN{1\textsuperscript{st} Mahdi Abolghasemi}
\IEEEauthorblockA{\textit{Department of Data Science and AI} \\
\textit{Monash University}\\
Melbourne, Australia \\
mahdi.abolghasemi@monash.edu}
\and
\IEEEauthorblockN{2\textsuperscript{nd} Rasul Esmaeilbeigi}
\IEEEauthorblockA{\textit{School of Information Technology} \\
\textit{Deakin University}\\
Melbourne, Australia \\
rasul.esmaeilbeigi@uon.edu.au
}
}

\maketitle

\begin{abstract}
    In this paper, we describe our proposed methodology to approach the ``predict+optimise'' challenge introduced in the IEEE CIS 3\textsuperscript{rd} Technical Challenge. The predictive model 
    employs an ensemble of LightGBM models and the prescriptive analysis employs mathematical optimisation to efficiently prescribe solutions that minimise the average cost over multiple scenarios. Our solutions ranked 1\textsuperscript{st} in the optimisation and 2\textsuperscript{nd} in the predict challenge of the competition.
\end{abstract}


\section{Introduction}\label{intro}
This paper explores the forecasting and optimisation models used for IEEE-CIS predict and optimise competition. The challenge was 1) to forecast fifteen minutely slots of power for six solar panels and six buildings and 2) to provide a schedule of batteries and activities such that the total energy cost is minimised. We started the forecasting part by conducting an extensive exploratory data analysis and looked at trends, seasonality, and intermittency pattern of data. We also explored the impact of COVID-19 on buildings power demand since part of the provided data and the forecasting horizon was during the pandemic. Since both solar power and buildings demand are highly dependent on weather conditions, we used the hourly weather data provided by the organisers of the competition and downloaded the daily data from the BOM website (www.bom.gov.au). We used various statistical and machine learning models including seasonal ARIMA, random forest (RF), LightGBM, and support vector regression (SVR) for building the predictive models. While the generated forecasts especially with RF and SVR were fairly accurate and competitive to LightGBM, we opt to use  LightGBM since it is significantly faster and returns reliable forecasts. The forecasting models were all trained with LightGBM where calendar features, daily weather data, hourly weather data and various rolling statistics of these features were used as input variables in the model. We optimised hyperparameters and selected the most significant features for each model. We generated several forecasts with different models to provide a larger pool of scenarios for the optimisation part of the competition. The final submitted forecasts were an ensemble of two LightGBM models for each series where daily and hourly features were used for each series and hyperparameters were optimised. 

The objective function of the optimisation part minimises the total energy cost that includes a quadratic form, the square of the maximum load. We used a linearisation technique to linearise this objective function and developed a mixed integer linear program that captures all constraints of the problem. 
One of the input parameters of the optimisation model is the \textit{net base load}, i.e., the total predicted base load of buildings minus generation of their solar panels, per time slot. 
Our proposed optimisation approach does not rely on one forecast. We consider the so-called Sample Average Approximation Method (SAAM) in which the optimisation model minimises the average cost of a solution over multiple scenarios (predictive outcomes) rather than just one. Our final submission employs 6 forecasting scenarios. This approach generally prescribes a solution with least expected cost that is also less sensitive to the forecasting errors. See \cite{esmaeilbeigi2021benders} for more details of SAAM. 

The rest of the paper is organised as follows. Section \ref{Background} briefly reviews the most relevant literature. Section \ref{Methodology} and \ref{Experiments} explain our methodology and details of experiments, respectively.

\section{Background}\label{Background}

Forecasting renewable energy and in particular solar power has been investigated extensively in the literature \cite{b2}. Solar power is highly uncertain and is directly impacted by weather conditions such as solar radiations, temperature, cloud coverage.  Many different forecasting methods have been proposed to forecast renewable energy. This includes machine learning, statistical models, physical models, and hybrid versions of them. There is no unique solution and superior foresting method, rather the performance of these methods depends on the forecasting horizon, series behaviour, granularity of series, and particularities of the problem \cite{b2}. Similarly forecasting the energy demand buildings have been studied extensively and various types of models have been used for forecasting energy demand in buildings. The literature suggests the performance of these models may depend on many factors such as ambient temperature, building operations like ventilation systems and lighting, and building structure among others \cite{b5}. While we can see a diverse set of models that are used in both domains, tree-based models and in particular LightGBM has been successfully implemented in various energy forecasting problems \cite{b5, b1}.  

Decision making under uncertainty plays a crucial role in managing energy systems. Uncertainty should be addressed properly since these systems are highly reliant on the predetermined energy prices and policies as well as predictable energy loads and demands \cite{dai2021utilization}. 
In the present paper, we consider the net base load of each time slot as a random variable in the optimisation model and employ SAAM to minimise the expected cost over some likely outcomes of this random variable.
\section{Methodology}\label{Methodology}
We started to develop predictive models by prepossessing the solar and buildings data.  There are many missing values in the data. At first, we aimed to replace the missing values with the average power across the same hours and days for each time slot but training the forecasting models on such data significantly reduced the accuracy of the forecasts. Therefore, we removed the missing values from the data to provide better quality data for training the models. 

Solar and building powers depict multiple seasonality patterns. Therefore, we created several calendar features such as minutes (15 minutely only), hour, day, week, day-of-week to capture the multiple seasonality behaviour inherent in solar data and buildings data. We gathered daily weather data including max temperature, min temperature, solar exposure, and rainfall from the BOM website and created various static and dynamic features such as lags, mean, standard deviation for the aforementioned daily information. This was done in a rolling window fashion. 
We also used the hourly data provided by the competition organisers that includes hourly weather data such as temperature, pressure, wind speed, humidity, surface thermal radiation, surface solar radiation,  total cloud coverage. We used the raw features and conducted a comprehensive feature engineering on this data to create dynamic features that can capture the stochastic behaviour of power over time. This includes lags of data up until lag three, mean, and standard deviation with the length of three to six hours. 

We developed three sets of models where we used daily weather data, hourly weather data, and a combination of daily and hourly weather data besides calendar features as input features of our models. The first attempt includes one LightGBM model for each series with the calendar and daily weather data as input features (12 models in total), and one LightGBM model for each series with the calendar and hourly weather features as input variables (12 models in total). We optimised the hyperparameters with grid search and trained 24 LightGBM models (one model with hourly features data and one model with daily features data for each series). We set the \textit{num-leaves} and \textit{min-data-in-leaf} between 50 to 300 with steps of 20,  \textit{max-depth} between 7 and 13 with a step of 1. All LightGBM models were trained with a learning rate equal to 0.1 and  \textit{mae} loss function.  We also examined the \textit{rmse} loss function but that significantly reduced the accuracy as predictions were evaluated based on ``mase".  In order to avoid overfitting, we used L1 regularisation with early stopping criteria. We conducted time series cross-validation where September and October data were used as evaluation sets and used L1 regularisation and early stopping to avoid overfitting. The length of the training set for each buildings and solars differs and they were selected as per our observations and experiments.  Our final submission was an ensemble of two LightGBM models for each series with daily and hourly features. The code is publicly available at \href{https://github.com/mahdiabolghasemi/IEEE-predict_optimise_technical_challenge}{GitHub}\footnote{\href{https://github.com/mahdiabolghasemi/IEEE-predict_optimise_technical_challenge}{github.com/mahdiabolghasemi/IEEE-predict-optimise-technical-challenge}}.


In terms of the importance of the variables, we can see calendar events and especially hour of the day consistently being ranked as an important feature for solar data. For building data, various features such as temperature, day, and hour were chosen as the top-ranked features depending on the series. We observe a similarity between the importance of features within buildings and solar data but their ranking differs.

We formulated the optimisation problem as a mixed-integer linear program in which the net base load is assumed to be a given deterministic parameter. 
The sets and parameters used to develop the optimisation model are defined as follows: $\mathcal{A}^R$ is the set of all recurring activities; $\mathcal{A}^O$ is the set of all once-off activities; $\mathcal{A}$ is the set of all activities ($\mathcal{A} \coloneqq \mathcal{A}^R \cup \mathcal{A}^O$); $\mathcal{B}$ is the set of all batteries; $\mathcal{T}$ is the set of all time slots in the planning horizon;  $\mathcal{T}_a \subset \mathcal{T}$ is the set of time slots when activity $a \in \mathcal{A}$ can be in progress. For $a \in \mathcal{A}^R$, this set only corresponds to the time slots of the first week. $\mathcal{T}'_a \subset \mathcal{T}_a$ is the set of time slots when activity $a \in \mathcal{A}$ can start. For $a \in \mathcal{A}^R$, this set only corresponds to the time slots of the first week.  $\mathcal{T}''_a \subset \mathcal{T}'_a$ is the set of starting time slots for activity $a \in \mathcal{A}^O$ that result in a penalty;  $\mathcal{P}_a$ is the set of prerequisites of activity $a \in \mathcal{A}$;  $p_a$ is the penalty of scheduling activity $a \in \mathcal{A}^O$ outside of working hours; $r_a$ is the revenue of scheduling activity $a \in \mathcal{A}^O$;  $\delta_a$ is the duration of activity $a \in \mathcal{A}$;  $n^S_a$ and $n^L_a$ are respectively the number of small and large rooms required for activity $a \in \mathcal{A}$; $\beta_a$    is the load required by activity $a\in\mathcal{A}$ per room; $S$ is the number of small rooms available in total; $L$  is the number of large rooms available in total; $c_b$ is the capacity of battery $b\in\mathcal{B}$; $c'_b$  is the energy of battery $b\in\mathcal{B}$ at the beginning of planning; $m_b$  is the maximum power of battery $b\in\mathcal{B}$; $e_b$  is the efficiency of battery $b\in\mathcal{B}$;  $\pi_t$ is the wholesale price at time $t \in \mathcal{T}$; $l_t$  is the net base load at time $t \in \mathcal{T}$; $M$ is an integer upper bound on the absolute value of the maximum load; $D$ is the number of time slots in a day; $T$ is the number of time slots in the planning horizon ($T\coloneqq|\mathcal{T}|$); $[t]$ is a function that maps $t \in \mathcal{T}$ to the corresponding time slot in the first week. 
    
The decision variables are defined as follows: $x_{bt}$ is a binary variable equal to 1 if and only if (iff) battery $b\in\mathcal{B}$ is charging at time $t \in \mathcal{T}$;  $y_{bt}$  is a binary variable equal to 1 iff battery $b\in\mathcal{B}$ is discharging at time $t \in \mathcal{T}$; $z_{at}$  is a binary variable equal to 1 iff activity $a\in\mathcal{A}$ begins at time $t \in \mathcal{T}'_a$; $v_{at}$   is a binary variable equal to 1 iff activity $a\in\mathcal{A}$ is in progress at time $t \in \mathcal{T}_a$; $s_{bt}$  is a non-negative variable representing the state of battery $b\in\mathcal{B}$ at the end of time slot $t \in \mathcal{T}$; $w_a$ is a binary variable equal to 1 iff activity $a\in\mathcal{A}$ is scheduled; $u_a$  is a binary variable equal to 1 iff activity $a\in\mathcal{A}^O$ is scheduled outside working hours;  $d_a$  is a non-negative variable representing the day index at which activity $a\in\mathcal{A}$ begins if it is scheduled; $\ell_t$  is an unrestricted variable representing the aggregate/total load at time $t \in \mathcal{T}$;    $\eta$  is a non-negative variable representing the absolute value of the maximum load;  $\lambda_i$  is a binary variable equal to 1 iff $\ceil{\eta}$ is equal to $i \in \{1,\dots,M\}$.

Objective \eqref{obj_deterministic} approximates the problem's quadratic cost function as a linear function. Constraints~\eqref{const_01}--\eqref{const_03} ensure that each activity starts in exactly one time slot and that the activity continues without any interruption until it finishes. 
According to Constraint~\eqref{const_04}, the auxiliary binary variables $u_a$ is equal to one iff activity $a \in \mathcal{A}^O$ is scheduled outside working hours. Constraints~\eqref{const_05}--\eqref{const_07} ensure that there is at least one day between start times of an activity and its prerequisites. Furthermore, an activity is not scheduled unless its prerequisites have all been scheduled. Constraints~\eqref{const_08}--\eqref{const_09} capture the capacity of each battery at a time slot depending on its initial state and the charging/discharging actions performed in that time slot. Constraint~\eqref{const_10} states that a battery cannot charge and discharge at the same time. Constraints~\eqref{const_12}--\eqref{const_13} guarantee that the total numbers of small and large rooms do not exceed their respective capacities. Constraint~\eqref{const_11} captures the value of the net load based on the predicted load, the load from the batteries and the load required by the scheduled activities. Constraints~\eqref{const_14}--\eqref{const_17} are used to capture the value of the maximum load $\eta$ and linearize the value of $\ceil{\eta}^2$ in conjunction with the objective function. This provides a reasonable approximation of the quadratic objective function. Constraint~\eqref{const_18} guarantees that recurring activities are all scheduled. Constraints~\eqref{const_19} ensures that the energy of a battery is always non-negative, and it does not exceed its capacity.

    {
        \footnotesize
    \begin{align}
        \min         \frac{0.25}{1000}\sum_{t \in \mathcal{T}}\pi_t\ell_t + 0.005\sum_{i=1}^{M} i^2\lambda_i - \sum_{a \in \mathcal{A}^O} (r_a w_a  -p_a u_a)\label{obj_deterministic}
    \end{align}
    \begin{align}
        & \sum_{t'\in\mathcal{T}'_a \cap \{t-\delta_a+1, \dots, t\}} z_{at'} = v_{at}                                                                                                    &  & a \in \mathcal{A}, t \in \mathcal{T}_a \label{const_01}                          \\
                     & \sum_{t \in \mathcal{T}_a} v_{at} = \delta_a w_a                                                                                               &  & a \in \mathcal{A} \label{const_02}                                               \\
                     & \sum_{t \in \mathcal{T}'_a} z_{at} = w_a                                                                                                       &  & a \in \mathcal{A} \label{const_03}                                               \\
                     & \sum_{t \in \mathcal{T}''_a} z_{at} = u_a                                                                                                      &  & a \in \mathcal{A}^O \label{const_04}                                             \\
                     & \sum_{t \in \mathcal{T}'_a} \bfdiv{t}{D} z_{at} + \bcdiv{T+1}{D}(1-w_a) = d_a                                                                                       &  & a \in \mathcal{A} \label{const_05}                                               \\
                     & d_a + w_a \le d_{a'}                                                                                                            &  & a' \in \mathcal{A}, a \in \mathcal{P}_{a'} \label{const_06}                      \\
                     & w_{a'} \le w_a                                                                                                                                 &  & a' \in \mathcal{A}, a \in \mathcal{P}_{a'} \label{const_07}                      \\
                     & s_{bt} = c'_b + 0.25\,m_b \left(x_{bt}-y_{bt}\right)                                                                                                 &  & b \in \mathcal{B}, t = 1 \label{const_08}                                        \\
                     &  s_{bt} = s_{b,t-1} + 0.25\,m_b \left(x_{bt}-y_{bt}\right)                                                                                            &  & b \in \mathcal{B}, t \in \mathcal{T}\setminus \{1\} \label{const_09} \\
                     & x_{bt}+y_{bt}\le 1                                                                                                                             &  & b \in \mathcal{B}, t \in \mathcal{T} \label{const_10}                            \\
                     & \sum_{a \in \mathcal{A}^O} n^L_a v_{at} + \sum_{a \in \mathcal{A}^R} n^L_a v_{a[t]} \le L                                                                                                    &  & t \in \mathcal{T} \label{const_12}                                               \\
                     & \sum_{a \in \mathcal{A}^O} n^S_a v_{at} + \sum_{a \in \mathcal{A}^R} n^S_a v_{a[t]} \le S                                                                                                    &  & t \in \mathcal{T} \label{const_13}                                               \\
                     & \ell_t = l_t + \sum_{b \in \mathcal{B}} \frac{m_b}{\sqrt{e_b}} \left(x_{bt}-e_b y_{bt}\right) \notag                                           &  &                                                                              \\
        {}           & ~~~~~~~~~ +\sum_{a \in \mathcal{A}^O} \beta_a (n^S_a+n^L_a) v_{at}                                                                                   &  & \notag                                               \\
        & ~~~~~~~~~ +\sum_{a \in \mathcal{A}^R} \beta_a (n^S_a+n^L_a) v_{a[t]}                                                                                    &  & t \in \mathcal{T} \label{const_11}                                               \\
                     & \sum_{i=1}^{M} \lambda_i \le 1                                                                                                                 &  & \label{const_14}                                                                 \\
                     & \sum_{i=1}^{M} i\lambda_i \ge \eta                                                                                                             &  & \label{const_15}                                                                 \\
                     & \eta \ge \ell_t                                                                                                                                &  & t \in \mathcal{T} \label{const_16}                                               \\
                     & \eta \ge -\ell_t                                                                                                                                &  & t \in \mathcal{T} \label{const_17}                                               \\
                     & w_a = 1                                                                                                                                        &  & a \in \mathcal{A}^R \label{const_18}                                             \\
                     & 0 \le s_{bt} \le c_b                                                                                                                           &  & b \in \mathcal{B}, t \in \mathcal{T} \label{const_19}
    \end{align}
    }
    The model does not consider the specific allocation of building rooms to activities to avoid symmetry and also reduce the problem size. This allocation is instead performed in a post-processing step, where a feasibility problem is modeled and solved. Due to the space limitation, this model is not presented here. The decision variables corresponding to the recurring activities are only defined for the first week to reduce the size of the formulation. These decision variables are used throughout the planning horizon by employing the mapping $[t]$. Some once-off activities will result in zero or negative revenue if they are scheduled outside of working hours. There exists an optimal solution in which these activities are not allocated to the time slots outside of the working hours. Therefore, we perform a pre-processing approach to avoid generating additional decision variables corresponding to such allocations.
    It is noteworthy to mention that Constraints~\eqref{const_05}--\eqref{const_06} adhere to the feasibility checks of the competition. We highlight that they can be generalized to capture prerequisites also for the case in which activities can be scheduled in the same day. This can be done by defining appropriate coefficients (see Section 6.3.2 of \cite{esmaeilbeigi2021multiphase} for details).

    
    The deterministic formulation assumes that the true realisation of the net base load is given ($l_t$ for $t \in \mathcal{T}$). We consider the uncertainty of forecasts within the optimisation problem and minimise the average cost over multiple uncertain scenarios of forecasts. Let $\mathcal{S}$ denote the set of scenarios and $l_{ts}$ be the realisation of random variable $l_t$ in scenario $s \in \mathcal{S}$. Accordingly, we define variables $\ell_{ts}$, $\eta_s$ and $\lambda_{is}$ corresponding to scenario $s \in \mathcal{S}$. The updated formulation has the objective function
    
    {
        \footnotesize
    \begin{align}
        \min~          & \frac{1}{|\mathcal{S}|}\sum_{s \in \mathcal{S}}\left(\frac{0.25}{1000}\sum_{t \in \mathcal{T}}\pi_t\ell_{ts} + 0.005\sum_{i=1}^{M} i^2\lambda_{is}\right) \notag \\
        {} - &\sum_{a \in \mathcal{A}^O} (r_a w_a  -p_a u_a)\label{obj_deterministic_2}
    \end{align}
    }
    which minimises the average cost incurred in different scenarios. Constraints \eqref{const_11}--\eqref{const_17} must hold in every scenario, and therefore they are respectively updated to

    {
        \footnotesize
    \begin{align}
        & \ell_{ts} = l_{ts} + \sum_{b \in \mathcal{B}} \frac{m_b}{\sqrt{e_b}} \left(x_{bt}-e_b y_{bt}\right) \notag                                           &  &                                                                              \\
        {}           & ~~~~~~~~~ +\sum_{a \in \mathcal{A}^O} \beta_a (n^S_a+n^L_a) v_{at}                                                                                   &  & \notag                                               \\
        & ~~~~~~~~~ +\sum_{a \in \mathcal{A}^R} \beta_a (n^S_a+n^L_a) v_{a[t]}                                                                                    &  & s \in \mathcal{S}, t \in \mathcal{T} \label{const_11_2}                                               \\
                     & \sum_{i=1}^{M} \lambda_{is} \le 1                                                                                                                 &  & s \in \mathcal{S} \label{const_14_2}                                                                 \\
                     & \sum_{i=1}^{M} i\lambda_{is} \ge \eta_s                                                                                                             &  & s \in \mathcal{S} \label{const_15_2}                                                                 \\
                     & \eta_s \ge \ell_{ts}                                                                                                                                &  & s \in \mathcal{S}, t \in \mathcal{T} \label{const_16_2}                                               \\
                     & \eta_s \ge -\ell_{ts}                                                                                                                                &  & s \in \mathcal{S}, t \in \mathcal{T} \label{const_17_2}
    \end{align}
    }
In order to solve the resulting mixed integer linear program, we employed Gurobi optimisation solver (version 9.1.2). We developed an optimiser engine in Python which is publicly available at \href{https://github.com/resmaeilbeigi/IEEE\_CIS\_3rd\_Technical\_Challenge\_Optimiser}{github}\footnote{\href{https://github.com/resmaeilbeigi/IEEE\_CIS\_3rd\_Technical\_Challenge\_Optimiser}{github.com/resmaeilbeigi/IEEE\_CIS\_3rd\_Technical\_Challenge\_Optimiser}}. Although we defined the auxiliary variables $\lambda_{i}$ or $\lambda_{is}$ as binary variables, we propose to consider them as non-negative continuous variables when solving the problem. This relaxation significantly improved the run time of the solver while providing a very good bound.


To further speed up solving the problem, we designed different fix-and-optimize heuristics to obtain good feasible solutions in a reasonable amount of time. A collection of 12 algorithms have been included in the optimiser engine. Due to the space limitation, we only explain the default algorithm here. The optimiser engine can start with a good initial feasible solution (provided by the user in the form of a solution file) to warm-start the solver. 
If this setting is activated (i.e., parameter \textsf{setstart} is set to \textsf{True}),
the engine will take a two-phase approach: in the first phase, the integrality constraints of all decision variables corresponding to the batteries are relaxed (e.g., allowing for a battery to be partially charged and discharged at the same time) while ensuring feasibility of the solution with regard to the activities. Once a solution for phase one is obtained, we fix the decision variables corresponding to the activities and optimise scheduling of the batteries by including integrality constraints of their decision variables. If the \textsf{setstart} parameter is \textsf{False}, the engine will take another two-phase approach as follows: In the first phase, no batteries and no penalized activities 
 are scheduled. Furthermore, the start time of all activities are restricted to the even time indices. In the second phase, we fix the activities that have been scheduled in phase one (with some degrees of flexibility)
 and schedule the batteries to improve the solution (peak-shaving).
For more details, please see the implementation of the optimiser engine.

\section{Experiments}\label{Experiments}

To solve the forecasting problem, we trained different LightGBM models where we used calendar features and various form of weather data as inputs. We initially set the hyper-parameters manually but then optimised them with grid search. We started by using  daily or hourly weather data as input features. We then developed a model with the combination of daily and hourly features which significantly improved the forecast accuracy. An ensemble of daily, hourly, and daily-hourly models resulted in the accuracy of 0.58 MASE over October testset. In our final submission, we did not train the combined daily-hourly model due to limitations in time.

To solve the optimisation problem, we explored various heuristics and examined various solver parameters. We gradually improved the algorithms by tuning these parameters and advancing the heuristics. The current default values of the settings in the optimiser engine proved to be promising. For the last submission, we employed the default algorithm and used solutions of the previous submission as the initial solution. 

In terms of the association between forecast accuracy and energy cost in our experiments, we observe that our submissions on November testset with accuracy of 0.84 and 0.74 respectively resulted in the actual energy costs of 337625 and 328359. This indicates a positive correlation between the accuracy of the forecast and the scheduling costs, where reducing MASE by 11\% resulted in 3\% reduction in scheduling costs in this case. Note that this correlation may not be linear and as accurate since the optimality gap of the solutions were nearly 2.5\%. Furthermore, such a comparison can be made on the basis of actual energy costs and not the costs reported by an optimisation algorithm.

\section*{CRediT author statement}
\textbf{Mahdi Abolghasemi}: Methodology, Validation, Analysis and Writing of Prediction part. \textbf{Rasul Esmaeilbeigi}: Methodology, Validation, Analysis and Writing of Optimisation part.

\vspace{12pt}
\color{red}


\begin{thebibliography}{00}
	
\bibitem{b1} Bergmeir, C., Abolghasemi, M., Vahid-Araghi, F., Hyndman, R.J, ``Renewable energy power forecasting - Very short term wind power case study''. Technical report for Australian Renewable Energy Agency, Monash University, 2020.

\bibitem{dai2021utilization}
{ Dai, R., Esmaeilbeigi, R., and Charkhgard, H.}
\newblock The utilization of shared energy storage in energy systems: A
 comprehensive review.
\newblock {\em IEEE Transactions on Smart Grid 12}, 4 (2021), 3163--3174.


\bibitem{esmaeilbeigi2021multiphase}
{ Esmaeilbeigi, R., Mak-Hau, V., Yearwood, J., and Vivian, N.}
\newblock The multiphase course timetabling problem.
\newblock {\em European Journal of Operational Research\/} (2021), 1--35.
\newblock https://doi.org/10.1016/j.ejor.2021.10.014

\bibitem{esmaeilbeigi2021benders}
{ Esmaeilbeigi, R., Middleton, R., Garc{\'\i}a-Flores, R., and Heydar, M.}
\newblock Benders decomposition for a reverse logistics network design problem
  in the dairy industry.
\newblock {\em Annals of Operations Research\/} (2021), 1--52.


\bibitem{b2} Hong, T., Pinson, P., Wang, Y., Weron, R., Yang, D. and Zareipour, H., ``Energy forecasting: A review and outlook.'' IEEE Open Access Journal of Power and Energy, 2020.



\bibitem{b5} Zhao, H.X. and Magoules, F., ``A review on the prediction of building energy consumption''. Renewable and Sustainable Energy Reviews, 16(6), pp.3586-3592, 2012.

\end{thebibliography}
\end{document}